\documentclass[sigconf]{acmart}

\usepackage{xcolor}
\usepackage{epsfig}
\usepackage{graphicx}
\usepackage{amsmath}
\usepackage{enumitem}
\usepackage{balance}

\usepackage[font=footnotesize,labelfont=bf]{caption}

\usepackage[belowskip=-5pt,aboveskip=3pt]{caption} 

\usepackage{multirow}
\usepackage{tabularx}
\usepackage{dsfont}
\usepackage{url}
\usepackage[tight]{subfigure}
\usepackage{color}
\usepackage{subfigure}
\usepackage{amsthm}
\usepackage[export]{adjustbox}
\usepackage{comment}
\usepackage[rightcaption]{sidecap}

\usepackage[utf8]{inputenc} 
\usepackage[T1]{fontenc}    
\usepackage{url}            
\usepackage{booktabs}       
\usepackage{amsfonts}       
\usepackage{nicefrac}       
\usepackage{microtype}      

\def\argmax{\operatornamewithlimits{arg\,max}}

\definecolor{dg}{rgb}{0,0.694,0.298}
\definecolor{purple}{rgb}{0.4,0.176,0.569}

\usepackage{pifont}
\usepackage{xspace}
\makeatletter
\DeclareRobustCommand\onedot{\futurelet\@let@token\@onedot}
\def\@onedot{\ifx\@let@token.\else.\null\fi\xspace}
\def\eg{\emph{e.g}\onedot} 
\def\ie{\emph{i.e}\onedot} 
 
\def\etc{\emph{etc}\onedot} 
 
\def\etal{\emph{et al}\onedot}
\makeatother

\AtBeginDocument{%
  \providecommand\BibTeX{{%
    \normalfont B\kern-0.5em{\scshape i\kern-0.25em b}\kern-0.8em\TeX}}}

\copyrightyear{2021}
\acmYear{2021}
\setcopyright{acmcopyright}\acmConference[MM '21]{Proceedings of the 29th ACM International Conference on Multimedia}{October 20--24, 2021}{Virtual Event, China}
\acmBooktitle{Proceedings of the 29th ACM International Conference on Multimedia (MM '21), October 20--24, 2021, Virtual Event, China}
\acmPrice{15.00}
\acmDOI{10.1145/3474085.3475171}
\acmISBN{978-1-4503-8651-7/21/10}

\begin{document}
\fancyhead{}
\title{\emph{AdvFilter}: Predictive Perturbation-aware Filtering against Adversarial Attack via Multi-domain Learning}

\author{Yihao Huang$^{1}$, \ Qing Guo$^{2\dagger}$, \ Felix Juefei-Xu$^{3}$, \ Lei Ma$^{4}$, \ Weikai Miao$^{1}$, Yang Liu$^{2,5}$, \ Geguang Pu$^{1\dagger}$}
\thanks{$^{\mathrm{\dagger}}$ Qing Guo and Geguang Pu are the corresponding authors (tsingqguo@ieee.org).}
\affiliation{\institution{
$^{1}$East China Normal University \country{China} \ \ $^{2}$Nanyang Technological University \country{Singapore}\\ }}
\affiliation{\institution{$^{3}$Alibaba Group \country{USA}\ $^{4}$University of Alberta \country{Canada}\ $^{5}$ Zhejiang Sci-Tech University, China}}



\renewcommand{\shortauthors}{Yihao Huang et al.}
\renewcommand{\authors}{Yihao Huang, Qing Guo, Felix Juefei-Xu, Lei Ma, Weikai Miao, Yang Liu, Geguang Pu}

\begin{abstract}
High-level representation-guided pixel denoising and adversarial training are independent solutions to enhance the robustness of CNNs against adversarial attacks by pre-processing input data and re-training models, respectively.
Most recently, adversarial training techniques have been widely studied and improved while the pixel denoising-based method is getting less attractive.
However, it is still questionable whether there exists a more advanced pixel denoising-based method and whether the combination of the two solutions benefits each other.
To this end, we first comprehensively investigate two kinds of pixel denoising methods for adversarial robustness enhancement (\ie, existing additive-based and unexplored filtering-based methods) under the loss functions of image-level and semantic-level, respectively, showing that \textit{pixel-wise filtering} can obtain much higher image quality (\eg, higher PSNR) as well as higher robustness (\eg, higher accuracy on adversarial examples) than existing pixel-wise additive-based method.
However, we also observe that the robustness results of the filtering-based method rely on the perturbation amplitude of adversarial examples used for training.
To address this problem, we propose \textit{predictive perturbation-aware~\&~pixel-wise filtering}, where \textit{dual-perturbation filtering} and an \textit{uncertainty-aware fusion} module are designed and employed to automatically perceive the perturbation amplitude during the training and testing process. The method is termed as \textbf{\emph{AdvFilter}}.
Moreover, we combine adversarial pixel denoising methods with three adversarial training-based methods, hinting that considering data and models jointly is able to achieve more robust CNNs.
The experiments conduct on NeurIPS-2017DEV, SVHN and CIFAR10 datasets and show advantages over enhancing CNNs' robustness, high generalization to different models and noise levels.

\end{abstract}





\maketitle

\section{Introduction}\label{sec:intro}
Due to the truth that neural networks are hard to explain, the security problem in deep learning tends to get a lot of attention. As one of the hot spots, the adversarial attack \cite{goodfellow2014explaining} is always the main method in invalidating the neural network. Thus how to make the neural network more robust is a really important problem.

In recent years, high-level representation-guided pixel denoising and adversarial training \cite{kurakin2016adversarial} are the most effective methods in enhancing the robustness of CNNs against adversarial attacks. Pixel denoising focuses on pre-processing the input data. With the strengthen of pixel denoising, the adversarial data examples are closer to the distribution of the training dataset of the CNNs. Furthermore, adversarial training allows the network to see more data distributions, which makes the CNNs more generalize.

As adversarial training has been widely studied, the pixel denoising-based method has been ignored. However, in our comprehensive investigation, we find that unexplored filtering-based pixel denoising does better than the existing additive-based pixel denoising under different loss functions (\eg, image-level and semantic-level). Furthermore, as pixel denoising and adversarial training enhance the CNNs from various perspectives, whether these two kinds of methods can benefit each other is an interesting problem. Thus we perform experiments on a basic adversarially trained model and two state-of-the-art adversarial training methods \cite{li2020enhancing,xie2019feature} to see the performance of pixel denoising to strengthen them.

Although filtering-based pixel denoising achieves better performance, it still has inherent limitations on dealing with adversarial noise images with different perturbation amplitudes (\ie, attack strengths). To address this problem, we propose a predictive perturbation-aware filtering method to automatically adapt to the amplitude of adversarial noise images.

Our denoising method not only maintains the advantage of the pure filtering-based method in classifying clean or low amplitude images but also gets better performance on images with median and large amplitude.

Experiments are conducted on three famous datasets NeurIPS-2017DEV \cite{kurakin2018adversarial}, SVHN \cite{svhn}, and CIFAR10 \cite{Krizhevsky09learningmultiple}.

The contributions can be summarized as follow.

\begin{itemize}

\item The first study on comparing the adversarial denoising defense performance of additive-based pixel denoising and filtering-based pixel denoising under different loss functions (\eg, image-level and semantic-level). 

\item We propose a predictive perturbation-aware filtering-based method to deal with adversarial attacks of varying attack strengths via multi-domain learning. The method robustifying the CNNs without harming accuracy on clean or tiny noise perturbation images.

\item We propose and verify an assumption that adversarial denoising and adversarial training can benefit each other on the adversarial defense task.

\end{itemize}

\section{Related Work}\label{sec:related}
{\bf Adversarial denoising defense.}
To our best knowledge, Liao \etal{} \cite{liao2018defense} is the only adversarial denoising defense method at present. For standard denoisers, small residual adversarial noises in the images are progressively amplified and lead to the wrong classification. To overcome this problem, Liao \etal{} \cite{liao2018defense} proposes high-level representation guided denoiser (HGD) as a defense for image classification. They take the difference of the target model’s (\eg, ResNet50 \cite{he2016deep}, VGG19 \cite{simonyan2014very}) intermediate representations (\ie, high-level representations) of clean image and denoised image to calculate the loss. However, HGD puts too much emphasis on high-level representation, its performance on tiny noise perturbation images is bad. Furthermore, HGD directly uses a neural network to predict the inverse noise map and adds it to the noise image for obtaining the denoise image. This additive-based denoising method is not as good as filtering-based denoising, which will be evaluated in Sec.~\ref{subsec:setups_empirical}.
\begin{figure}
\centering
\includegraphics[width=\linewidth]{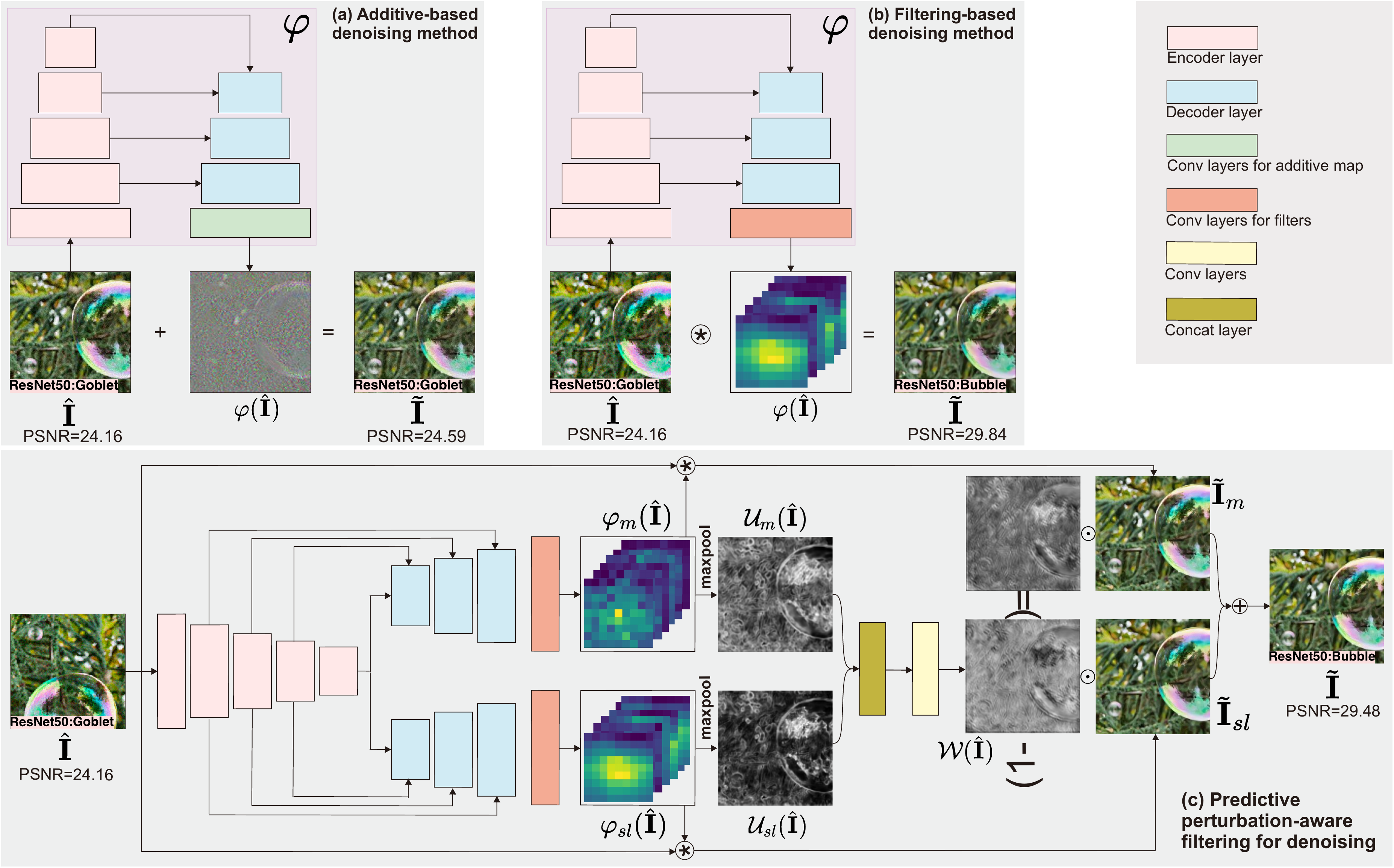}
\caption{Architectures of additive-based pixel denoising (\ie, (a)), filtering-based pixel denoising (\ie, (b)), and the proposed predictive perturbation-aware filtering (\ie, (c)) for adversarial robustness enhancement.}
\label{fig:archs}
\vspace{-10pt}
\end{figure}

{\bf Adversarial training defense.}
Adversarial training defense \cite{goodfellow2014explaining,kurakin2016adversarial,tramer2017ensemble,madry2017towards} is one of the most extensively investigated defenses against adversarial attacks. 

Yan \etal{} \cite{yan2018deep} integrate an adversarial perturbation-based regularizer into the classification objective, such that the obtained models learn to resist potential attacks. Xie \etal{} \cite{xie2019feature} develop a new network which contains blocks that performing feature denoising to increase adversarial robustness by performing feature denoising. Li \etal{} \cite{li2020enhancing} proposed feature pyramid decoder (FPD) framework implants a target CNN with both denoising and image restoration modules to filter an input image at multiple levels.

Adversarial training could achieve good results in defending against white-box and black-box attacks. However, it requires to involve a sufficient amount and variety of adversarial samples, leading to a high time complexity with huge GPU computing resources.

{\bf Other adversarial defense methods.}
There are some other adversarial defense methods. Das \etal{} \cite{das2017keeping} preprocess images with JPEG compression to reduce the effect of adversarial noises. Guo \etal{} \cite{guo2017countering} countering adversarial images using input transformations. Efros \etal{} \cite{efros2001image} uses image quilting for texture synthesis and transfer to response adversarial attacks.
Although these methods seem to be useful, Athalye \etal{} \cite{athalye2018obfuscated} shows that these defense methods significantly relied on gradient masking, which leads to a false sense of robustness against adversarial attacks.

{\bf General image denoising}
There are many general image denoising methods \cite{deng1993adaptive,rudin1992nonlinear,eng2001noise,mildenhall2018burst} that are intensively studied in image processing and computer vision. They commonly use filters to denoise images. Rudin \etal{} \cite{rudin1992nonlinear} propose a constrained optimization type of numerical algorithm for removing noise from images. 
Nowadays, Mildenhall \etal{} \cite{mildenhall2018burst} uses a kernel prediction network to build a filter variant method, which provides a specialized denoise filter for each pixel in the image.

\section{Pixel Denoising against Adversarial Attack}\label{sec:denoising}

Liao \etal \cite{liao2018defense} proposed the high-level representation guided denoising against adversarial attack, training an additive-based denoiser (\ie, adding predictive inverse noise map to the adversarial example for denoising) with the loss functions of image-level or semantic-level between clean and denoised images.
However, from the angle of image denoising, image filtering is a more advanced technique. For example, BM3D \cite{dabov2007image} method is the classical filtering-based denoising method that uses priors and non-linear optimization to recover a signal from a noisy image. Moreover, kernel prediction network-based denoising \cite{mildenhall2018burst} estimates pixel-wise kernels to filter each pixel and realizes state-of-the-art denoising results. 
Nevertheless, the effectiveness of such a more advanced denoising technique against adversarial attacks is still unknown.

In this section, we comprehensively study the effectiveness of the two pixel denoising methods under different loss function setups to see whether the filtering-based method is better than the additive-based method and to explore the limitations of the filtering-based denoiser for adversarial robustness enhancement. 
\subsection{Formulations}\label{subsec:formulations}
Given a standard trained CNN (\ie, $\phi(\cdot)$) for image classification and a clean image $\mathbf{I}$, we can generate imperceptible adversarial example $\hat{\mathbf{I}}$ via adversarial attack methods. In general, we use the following process for untargeted attack
%
\begin{align}\label{eq:adv}
\hat{\mathbf{I}} = \mathbf{I}+\delta,
\end{align}
%
where $\delta$ is the adversarial perturbation and obtained via
%
\begin{align}\label{eq:pgd}
\delta = \argmax_{\delta^{*}}{\mathcal{J}(\phi(\mathbf{I}+\delta^{*}),y),~\text{subject to}~\|\delta^{*}\|_p\leq\epsilon},
\end{align}
%
where $\mathcal{J}(\cdot)$ denotes the cross-entropy loss function with $y$ being the annotation of the input image.
In this paper, we employ the widely used projected gradient descent \cite{madry2017towards} (PGD) attack to solve Eq.~\eqref{eq:pgd}.
The adversarial examples can mislead the state-of-the-art CNNs, leading to robustness and security issues. 

Given an adversarial example (\eg, $\hat{\mathbf{I}}$), we aim to enhance CNNs' robustness by removing the adversarial perturbation through a denoising operation $\mathcal{D}(\cdot)$ with the predictive noise information.
Here, we use a pre-trained deep model $\varphi(\cdot)$ to predict the noise-related information directly and formulate the above process as
%
\begin{align}\label{eq:unifieddenoiser}
\tilde{\mathbf{I}} = \mathcal{D}(\varphi(\hat{\mathbf{I}}),\hat{\mathbf{I}}),
\end{align}
%
where $\tilde{\mathbf{I}}$ denotes the denoised image. To train the $\varphi(\cdot)$, we consider two kinds of loss functions, \ie, image-level and semantic-level difference between clean and denoised images as done in \cite{liao2018defense}
%
\begin{align}\label{eq:loss_img}
\mathcal{L}_\text{img} = \|\mathbf{I}-\tilde{\mathbf{I}}\|_1,
\end{align}
%
\begin{align}\label{eq:loss_sem}
\mathcal{L}_\text{sem} = \|\phi_l(\mathbf{I})-\phi_l(\tilde{\mathbf{I}})\|_1,
\end{align}
%
where $\phi_l(\cdot)$ denotes the output of the $l$th layer in the CNN $\phi(\cdot)$. The Eq.~(\ref{eq:loss_img}) encourages the denoised image to have the same image quality with the clean counterpart while the Eq.~(\ref{eq:loss_sem}) makes the deep features of $\mathbf{I}$ and $\tilde{\mathbf{I}}$ to be consistent, that is, the CNN $\phi(\cdot)$ predicts the same labels on $\mathbf{I}$ and $\tilde{\mathbf{I}}$. When setting different kinds of $\mathcal{D}(\cdot)$, we can get the additive-based and filtering-based pixel denoisers. 

{\bf Additive-based pixel denoising.} We get the additive-based pixel denoising by specifying Eq.~\eqref{eq:unifieddenoiser} as 
%
\begin{align}\label{eq:adddenoiser}
\tilde{\mathbf{I}} = \mathcal{D}(\varphi(\hat{\mathbf{I}}),\hat{\mathbf{I}})=\hat{\mathbf{I}}+\varphi(\hat{\mathbf{I}}),
\end{align}
%
where the deep model $\varphi(\hat{\mathbf{I}})$ predicts the inverse adversarial perturbation, \ie, $-\delta$ in Eq.~\eqref{eq:adv}, and the additive operation is used to perform the denoising.

{\bf Filtering-based pixel denoising.} We realize the filtering-based pixel denoising by specifying Eq.~\eqref{eq:unifieddenoiser} as 
%
\begin{align}\label{eq:filterdenoiser}
\tilde{\mathbf{I}} = \mathcal{D}(\varphi(\hat{\mathbf{I}}),\hat{\mathbf{I}})=\hat{\mathbf{I}}\circledast\varphi(\hat{\mathbf{I}}),
\end{align}
%
where `$\circledast$' is the pixel-wise filtering, that is, each pixel of $\hat{\mathbf{I}}$ is filtered by an exclusive kernel (of which the kernel size is $K$) estimated from $\varphi(\hat{\mathbf{I}})\in\mathbb{R}^{H\times W\times 3K^2}$. Specifically, for $p$th pixel of $\hat{\mathbf{I}}$, we filter it via an exclusive kernel with size of $K\times K\times 3$ that are stored at the $p$th location of $\varphi(\hat{\mathbf{I}})$.
We show their architectures in Fig.~\ref{fig:archs}.

\subsection{Setups for Empirical Study}\label{subsec:setups_empirical}
In this part, we detail the experimental setups for the subsequent analysis and discussion experiments.

{\bf Adversarial attack method.} Following the protocol of \cite{xie2019feature} and \cite{Kannan2018NIPS}, we use the projected gradient descent (PGD) with $L_\infty$ implemented by Foolbox \cite{rauber2017foolbox} as the white-box attack method that is an iterative attack. There are two key hyper-parameters of the PGD method: the maximum perturbation for each pixel (\ie, $\epsilon$ in Eq.~\eqref{eq:pgd}) that is also known as attack strength, the number of attack iterations $n$, both of which affects the attack results significantly. We denote a PGD attack with $n$ and $\epsilon$ as $\text{PGD}(n,\epsilon)$. 

{\bf Threat model.} We select the ResNet50 \cite{He2016} and VGG16 \cite{simonyan2014very} as the threat model for subsequent empirical study.

{\bf Training Dataset.} 
We randomly choose 5 images from each category in the training dataset of ImageNet \cite{deng2009imagenet} and get 5,000 images. Then, for each threat model, we use $\text{PGD}(n=40,\epsilon)$ on the threat model to generate adversarial examples of each image under 12 different attack strength, \ie, $\epsilon\in\{1\mathrm{e}^{-m}, 3\mathrm{e}^{-m}, 5\mathrm{e}^{-m}|m\in[1,2,3,4]\}$ for the intensity values of image ranging from 0 to 1. The reason 
to use n=40 is that 40 is the default parameter of Foolbox and can obtain a high attack success rate. Finally, we get a total of 60,000 adversarial examples for 5,000 clean images, that is, we have 60,000 image pairs as the training dataset of the denoising models.

{\bf Testing Dataset.} We use the NeurIPS-2017 DEV dataset \cite{kurakin2018adversarial} as the testing dataset, which is compatible with the ImageNet dataset and used by NIPS 2017 adversarial attacks and defenses competition. We also use the CIFAR10 \cite{Krizhevsky09learningmultiple} and SVHN \cite{svhn} datasets as the testing dataset. Similar to the training dataset, we perform $\text{PGD}(n,\epsilon)$ attack on the images with different $n$ and $\epsilon$. For each $\epsilon$ and each dataset, 1000 images are used for testing. As a result, we can evaluate the accuracy of models under different kinds of PGD attacks.

{\bf Architecture of $\varphi(\cdot)$.} For a fair comparison, we choose the widely used U-Net \cite{ronneberger2015u} as the $\varphi(\cdot)$ for both additive-based and filtering-based denoising methods. The U-Net has eight blocks. Each block has three convolutional layers with a kernel size of three and ReLU activation functions. The first five blocks mainly extract the features of the image and generate an intermediate representation of shape 14 $\times$ 14 with 512 channels while the last three blocks make up the decoder. Blocks 2,3,4 have skip connections to the last three blocks respectively. The main difference between additive-based and filtering-based method is the final output layer which behind the eight blocks. For the additive-based pixel denoising (\ie,Eq.~\eqref{eq:adddenoiser}), $\varphi(\hat{\mathbf{I}})\in\mathbb{R}^{H\times W\times 3}$ has the same size with $\hat{\mathbf{I}}\in\mathbb{R}^{H\times W\times 3}$ for pixel-wise additive operation in Eq.~\eqref{eq:adddenoiser}. For the filtering-based method (\ie, Eq.~\eqref{eq:filterdenoiser}), $\varphi(\hat{\mathbf{I}})\in\mathbb{R}^{H\times W\times 3K^2}$ contains the kernels for each pixel.

\begin{SCtable*}
\scriptsize
\centering
\caption{Accuracy of the standard trained networks (\eg, ResNet50, VGG16) on adversarial examples of datasets (\eg, NeurIPS-2017DEV, CIFAR10, SVHN) (\ie, Acc. after Attack), and their denoised counterparts with four additive \& filtering-based denoising methods.}
\setlength{\tabcolsep}{5pt} 
\begin{tabular}{llccccccccccccc}
\midrule
& \multicolumn{13}{c}{$\text{PGD}(n=40,\epsilon)$} \tabularnewline
& & 0 & $1\mathrm{e}^{-4}$ & $3\mathrm{e}^{-4}$ & $5\mathrm{e}^{-4}$ & $1\mathrm{e}^{-3}$ & $3\mathrm{e}^{-3}$ & $5\mathrm{e}^{-3}$ & $1\mathrm{e}^{-2}$ & $3\mathrm{e}^{-2}$ & $5\mathrm{e}^{-2}$ & $1\mathrm{e}^{-1}$ & $3\mathrm{e}^{-1}$ & $5\mathrm{e}^{-1}$ \tabularnewline
\midrule
Network \& Dataset & Acc. after Attack & 0.913 & 0.866 & 0.733 & 0.605 & 0.268 & 0.018 & 0.006 & 0 & 0 & 0 & 0 & 0 & 0 \tabularnewline
\midrule
\multirow{5}{*}{ResNet50 \& NeurIPS-2017DEV} &
Add~($L_\text{sem}$) & 0.906 & 0.885 & 0.825 & 0.740 & 0.532 & 0.083 & 0.026 & 0.002 & 0 & 0 & 0.001 & 0.001 & 0.001\tabularnewline 
& Filt($L_\text{sem}$) & 0.909 & 0.887 & 0.829 & 0.751 & 0.549 & 0.085 & 0.026 & 0.004 & 0.003 & 0.005 & 0.034 & 0.161 & 0.103\tabularnewline
& Add~($L_\text{img}$) & 0.914 & 0.872 & 0.749 & 0.625 & 0.295 & 0.020 & 0.007 & 0 & 0 & 0 & 0 & 0.005 & 0.001\tabularnewline
& Filt($L_\text{img}$) & 0.928 & 0.899 & 0.829 & 0.747 & 0.484 & 0.045 & 0.014 & 0.004 & 0.031 & 0.076 & 0.108 & 0.278 & 0.036\tabularnewline
& Filt($L_\text{img}^{*}$) & 0.791 & 0.791 & 0.783 & 0.776 & 0.758 & 0.694 & 0.634 & 0.525 & 0.298 & 0.168 & 0.048 & 0.001 & 0.000\tabularnewline
\midrule 
\midrule
& Acc. after Attack & 0.847 & 0.803 & 0.657 & 0.535 & 0.254 & 0.026 & 0.007 & 0.004 & 0.002 & 0.001 & 0.001 & 0 & 0 \tabularnewline
\midrule
\multirow{4}{*}{VGG16 \& NeurIPS-2017DEV} &
Add~($L_\text{sem}$) & 0.842 & 0.815 & 0.736 & 0.644 & 0.415 & 0.055 & 0.015 & 0.005 & 0.002 & 0.002 & 0.002 & 0.001 & 0.001\tabularnewline
& Filt($L_\text{sem}$) & 0.842 & 0.815 & 0.737 & 0.644 & 0.446 & 0.073 & 0.017 & 0.006 & 0.002 & 0.002 & 0.001 & 0 & 0\tabularnewline
& Add~($L_\text{img}$) & 0.846 & 0.803 & 0.657 & 0.535 & 0.254 & 0.026 & 0.007 & 0.004 & 0.002 & 0.001 & 0.001 & 0 & 0\tabularnewline
& Filt($L_\text{img}$) & 0.843 & 0.818 & 0.748 & 0.653 & 0.439 & 0.065 & 0.017 & 0.006 & 0.004 & 0.006 & 0.043 & 0.140 & 0.037\tabularnewline
\midrule 
\midrule
& Acc. after Attack & 0.918 & 0.927 & 0.905 & 0.869 & 0.833 & 0.603 & 0.345 & 0.075 & 0 & 0 & 0 & 0 & 0\tabularnewline
\midrule
\multirow{4}{*}{ResNet50 \& CIFAR10} &
Add~($L_\text{sem}$) & 0.921 & 0.929 & 0.908 & 0.87 & 0.833 & 0.609 & 0.349 & 0.076 & 0 & 0 & 0 & 0 & 0\tabularnewline
& Filt($L_\text{sem}$) & 0.907 & 0.915 & 0.902 & 0.883 & 0.868 & 0.783 & 0.660 & 0.423 & 0.106 & 0.051 & 0.021 & 0.011 & 0.002\tabularnewline
& Add~($L_\text{img}$) & 0.918 & 0.927 & 0.905 & 0.869 & 0.833 & 0.603 & 0.345 & 0.075 & 0 & 0 & 0 & 0 & 0\tabularnewline
& Filt($L_\text{img}$) & 0.900 & 0.907 & 0.887 & 0.880 & 0.876 & 0.787 & 0.699 & 0.437 & 0.084 & 0.029 & 0.007 & 0.001 & 0\tabularnewline
\midrule 
\midrule
& Acc. after Attack & 0.925 & 0.936 & 0.905 & 0.918 & 0.891 & 0.781 & 0.647 & 0.338 & 0.040 & 0.008 & 0.005 & 0.001 & 0\tabularnewline
\midrule
\multirow{4}{*}{ResNet50 \& SVHN} &
Add~($L_\text{sem}$) & 0.891 & 0.903 & 0.881 & 0.890 & 0.857 & 0.82 & 0.745 & 0.511 & 0.200 & 0.123 & 0.094 & 0.113 & 0.096\tabularnewline
& Filt($L_\text{sem}$) & 0.911 & 0.924 & 0.893 & 0.913 & 0.914 & 0.885 & 0.907 & 0.869 & 0.756 & 0.646 & 0.502 & 0.208 & 0.104\tabularnewline
& Add~($L_\text{img}$) & 0.925 & 0.935 & 0.905 & 0.920 & 0.892 & 0.781 & 0.650 & 0.345 & 0.041 & 0.008 & 0.005 & 0.001 & 0\tabularnewline
& Filt($L_\text{img}$) & 0.925 & 0.932 & 0.910 & 0.931 & 0.904 & 0.852 & 0.787 & 0.553 & 0.149 & 0.054 & 0.006 & 0.001 & 0\tabularnewline 
\midrule 

\end{tabular}
\label{Table:Compare_with_Add_Filt}
\end{SCtable*}

\subsection{Additive-based vs. filtering-based denoising}

\begin{SCtable*}
\scriptsize
\centering
\caption{Image quality enhancement by using additive \& filtering-based denoising methods (\ie, the group `ResNet50 \& NeurIPS-2017DEV' in Table~\ref{Table:Compare_with_Add_Filt}).}
\setlength{\tabcolsep}{4.5pt} 
\begin{tabular}{llccccccccccccc}
\midrule
& \multicolumn{13}{c}{$\text{PGD}(n=40,\epsilon)$} \tabularnewline
& & 0 & $1\mathrm{e}^{-4}$ & $3\mathrm{e}^{-4}$ & $5\mathrm{e}^{-4}$ & $1\mathrm{e}^{-3}$ & $3\mathrm{e}^{-3}$ & $5\mathrm{e}^{-3}$ & $1\mathrm{e}^{-2}$ & $3\mathrm{e}^{-2}$ & $5\mathrm{e}^{-2}$ & $1\mathrm{e}^{-1}$ & $3\mathrm{e}^{-1}$ & $5\mathrm{e}^{-1}$ \tabularnewline
\midrule

\multirow{2}{*}{PSNR} & Add($L_\text{sem}$) & 30.130 & 30.130 & 30.129 & 30.129 & 30.126 & 30.102 & 30.057 & 29.877 & 28.569 & 26.933 & 23.046 & 14.256 & 10.590\tabularnewline
&Filt($L_\text{sem}$) & 29.893 & 29.893 & 29.893 & 29.892 & 29.890 & 29.867 & 29.826 & 29.670 & 28.811 & 27.950 & 26.277 & 22.532 & 19.232\tabularnewline

\multirow{2}{*}{SSIM} &Add($L_\text{sem}$) & 0.959 & 0.959 & 0.959 & 0.959 & 0.959 & 0.958 & 0.957 & 0.953 & 0.919 & 0.863 & 0.692 & 0.298 & 0.166 \tabularnewline
&Filt($L_\text{sem}$) & 0.958 & 0.958 & 0.958 & 0.958 & 0.958 & 0.957 & 0.957 & 0.953 & 0.934 & 0.913 & 0.868 & 0.723 & 0.565 \tabularnewline

\multirow{2}{*}{PSNR} & Add($L_\text{img}$) & 31.765 & 31.765 & 31.764 & 31.763 & 31.759 & 31.722 & 31.658 & 31.417 & 29.946 & 28.234 & 24.404 & 16.499 & 13.511\tabularnewline
&Filt($L_\text{img}$) & 32.375 & 32.375 & 32.375 & 32.374 & 32.370 & 32.340 & 32.289 & 32.116 & 31.594 & 30.926 & 28.310 & 22.006 & 20.770\tabularnewline

\multirow{2}{*}{SSIM} &Add($L_\text{img}$) & 0.971 & 0.971 & 0.971 & 0.971 & 0.971 & 0.971 & 0.970 & 0.966 & 0.937 & 0.888 & 0.739 & 0.339 & 0.220\tabularnewline
&Filt($L_\text{img}$) & 0.975 & 0.975 & 0.975 & 0.975 & 0.975 & 0.975 & 0.974 & 0.972 & 0.964 & 0.953 & 0.913 & 0.752 & 0.614\tabularnewline

\midrule 
\end{tabular}
\label{Table:Filtering_additive_similarity}
\end{SCtable*}

\begin{SCtable*}
\scriptsize
\centering
\caption{Accuracy of the standard trained ResNet50 on adversarial examples of NeurIPS-2017DEV dataset (\ie, Acc. after Attack), and their denoised counterparts with multi-head filtering-based denoising method.}
\setlength{\tabcolsep}{4.5pt}
\begin{tabular}{ccccccccccccccc}
\hline 
& \multicolumn{13}{c}{$\text{PGD}(n=40,\epsilon)$}\tabularnewline
& & 0 & $1\mathrm{e}^{-4}$ & $3\mathrm{e}^{-4}$ & $5\mathrm{e}^{-4}$ & $1\mathrm{e}^{-3}$ & $3\mathrm{e}^{-3}$ & $5\mathrm{e}^{-3}$ & $1\mathrm{e}^{-2}$ & $3\mathrm{e}^{-2}$ & $5\mathrm{e}^{-2}$ & $1\mathrm{e}^{-1}$ & $3\mathrm{e}^{-1}$ & $5\mathrm{e}^{-1}$ \tabularnewline
\midrule
\multicolumn{2}{c}{Acc. after Attack}  & 0.913 & 0.886 & 0.733 & 0.605 & 0.268 & 0.018 & 0.006 & 0 & 0 & 0 & 0 & 0 & 0\tabularnewline
\midrule
\multirow{4}{*}{Multi-head-Filt($L_\text{img}$)} & $\text{head}\_1$ & 0.914 & 0.869 & 0.750 & 0.631 & 0.285 & 0.020 & 0.006 & 0 & 0 & 0 & 0 & 0 & 0\tabularnewline
& $\text{head}\_2$ & 0.914 & 0.869 & 0.752 & 0.633 & 0.287 & 0.020 & 0.006 & 0 & 0 & 0 & 0 & 0 & 0\tabularnewline
& $\text{head}\_3$ & 0.916 & 0.879 & 0.798 & 0.694 & 0.375 & 0.032 & 0.006 & 0 & 0 & 0 & 0 & 0 & 0\tabularnewline
& $\text{head}\_4$ & 0.792 & 0.788 & 0.783 & 0.778 & 0.759 & 0.694 & 0.635 & 0.527 & 0.297 & 0.163 & 0.047 & 0.001 & 0 \tabularnewline
\hline 
\end{tabular}
\label{Table:Multi-head-study}
\end{SCtable*}

According to the setups in Sec.~\ref{subsec:setups_empirical}, we conduct comprehensively empirical study to compare the existing additive-based denoising (\ie, `Add' in Table~\ref{Table:Compare_with_Add_Filt} ) and unexplored filtering-based denoising (\ie, `Filt' in Table~\ref{Table:Compare_with_Add_Filt}) for adversarial robustness enhancement under the two loss functions in Eq.~(\ref{eq:loss_img}) and (\ref{eq:loss_sem}). In Table~\ref{Table:Compare_with_Add_Filt} and ~\ref{Table:Filtering_additive_similarity}, Add($L_\text{img}$) and Filt($L_\text{img}$) represent the additive-based and filtering-based denoising method with image-level loss respectively while Add($L_\text{sem}$) and Filt($L_\text{sem}$) means denoising method with semantic-level loss. In table~\ref{Table:Compare_with_Add_Filt}, there are four groups of `Network \& Dataset'. We take the first group `ResNet50 \& NeurIPS-2017DEV' as an example to introduce our findings. The results in the following three groups show the generalization of our conclusion.

{\bf Training denoising methods.} With the same training strategy and dataset, we train additive and filtering-based denoisers under two loss functions (\ie, Eq.~\eqref{eq:loss_img}, \eqref{eq:loss_sem}), respectively, and get four categories (\ie, Add~($L_\text{sem}$), Filt($L_\text{sem}$), Add~($L_\text{img}$), Filt($L_\text{img}$)) whose names are shown in the second column of Table~\ref{Table:Compare_with_Add_Filt}.

{\bf Comparison results under a wide range of attack strengths.}
We calculate the classification accuracy of the threat model on the testing dataset. (\ie, the adversarial examples generated by $\text{PGD}(n=40,\epsilon)$ with $\epsilon$ from $0$ to $0.5$ where $\epsilon=0$ represents the clean images.)
Then, we use the four denoisers to pre-process all images and evaluate the new accuracy on the threat model to compare the effectiveness of different denoisers.
The higher accuracy after denoising indicates better denoisers against adversarial attacks.
According to the results of Table~\ref{Table:Compare_with_Add_Filt}, we observe that: \ding{182} The attack $\text{PGD}(n=40,\epsilon)$ leads to significant accuracy reduction as the attack strength, \ie, $\epsilon$, becomes larger (see the row `Acc. after Attack' in Table~\ref{Table:Compare_with_Add_Filt}). In particular, even with the very small $\epsilon=3\mathrm{e}^{-3}$, the accuracy is reduced from 91.3\% to 26.8\%, demonstrating the adversarial attack indeed affects the accuracy of standard trained CNN significantly. 
\ding{183} Compared with the accuracy after the attack, all denoisers show the capability of enhancing robustness under different attack strengths. Specifically, comparing `Add' with `Filt', we see that the filtering-based method shows much higher accuracy than the additive-based method under two loss functions on most attack strengths, demonstrating that the \textit{filtering-based denoising method is more effective than the additive-based method for adversarial robustness enhancement.}
\ding{184} Comparing different loss functions under the filtering-based method, we observe that the image-level loss function (\ie, $L_\text{img}$) shows significant advantages over the other semantic loss functions on the small strengths (\ie,  $\epsilon=[0,1\mathrm{e}^{-4},3\mathrm{e}^{-4}]$) and large attack strengths (\ie,  $\epsilon=[1\mathrm{e}^{-2},3\mathrm{e}^{-2},5\mathrm{e}^{-2},1\mathrm{e}^{-1},3\mathrm{e}^{-1}]$). In particular, when addressing the clean images ($\epsilon=0$), only denoising method with $L_\text{img}$ increase the accuracy further. 
These properties make the filtering-based method with image-level loss function a good candidate for adversarial robustness enhancement.

Nevertheless, we also note that Filt($L_\text{img}$) becomes less effective on the median attack strengths, which has lower accuracy than the methods with $L_\text{sem}$.
We suspect these results are related to the perturbation-aware property of Filt($L_\text{img}$) introduced by the training dataset.
Specifically, the training dataset contains the adversarial examples with all twelve strengths and we desire the $\varphi(\cdot)$ to predict proper pixel-wise filters for all strengths.
Intuitively, it is easy to estimate proper filters for the small and large attack strengths since they have significantly different patterns. However, it is hard to decide pixel-wise filters for the median strengths since their patterns are not so different from the small or large ones.
To further validate this point, we train the Filt($L_\text{img}$) with the adversarial examples generated by PGD with superior strengths and denotes it as Filt($L_\text{img}^{*}$). As shown in the last row of the group `ResNet50 \& NeurIPS-2017DEV' of Table~\ref{Table:Compare_with_Add_Filt}, we see the Filt($L_\text{img}^{*}$) achieves much higher accuracy on the median strengths than other denoisers.
We actually desire a denoiser that could combine the advantages of Filt($L_\text{img}$) and Filt($L_\text{img}^{*}$) and achieve high accuracy across all attack strength, hinting a perturbation-aware denoiser.

{\bf Image quality comparison.} Pixel-wise filtering-based method is not only more robust than additive-based method but also obtains higher image quality (\eg, higher peak signal-to-noise ratio (PSNR) \& structural similarity index measure (SSIM) \cite{hore2010image}). As shown in Table~\ref{Table:Filtering_additive_similarity}, we see that the filtering-based method has higher quality than the additive-based method on the image quality enhancement among most of the 13 attack strengths and two different loss functions. In particular, among the four types of denoising methods, Filt($L_\text{img}$) achieves the best image quality.

Overall, according to the empirical study, we conclude that: \ding{182} filtering-based denoising method shows absolute advantages for adversarial robustness enhancement over the existing additive-based denoising under both image-level and semantic-level loss functions. \ding{183} the filtering-based denoiser with the image-level loss function can achieve higher accuracy on both small and large attack strengths but is less effective on the median attack strengths, hinting us to develop a perturbation-aware denoiser.


\subsection{Multi-head filtering-based denoising}\label{subsec:multi_head_filtering}
In order to propose a perturbation-aware filtering-based denoising method, we first train a multi-head filtering-based denoising model to evaluate the performance of the filters trained with images of different attack strengths. 

As shown in Table~\ref{Table:Multi-head-study}, we have trained a model with four heads. The model has the same U-Net and each head has the same architecture as the final output layer of the filtering-based method introduced in Sec.~\ref{subsec:setups_empirical}. In the training procedure of the model, each head (\ie, \{$\text{head}\_{i}|i\in[1,2,3,4]\}$) is updated when meeting PGD-attacked images under three corresponding different attack strength (\ie, $\epsilon\in\{1\mathrm{e}^{i-5}, 3\mathrm{e}^{i-5}, 5\mathrm{e}^{i-5}\}$). 

We observe that: \ding{182} the head trained with the specific attack strengths may not has the best performance on the corresponding adversarial attacked images. For example, $\text{head}\_1$ is trained with images of adversarial strength $\epsilon\in\{1\mathrm{e}^{-4}, 3\mathrm{e}^{-4}, 5\mathrm{e}^{-4}\}$. However, the head which has the best denoising effect on $\epsilon\in\{1\mathrm{e}^{-4}, 3\mathrm{e}^{-4}\}$ is $\text{head}\_3$ while the best head for $\epsilon\in\{5\mathrm{e}^{-4}\}$ is $\text{head}\_4$. This means that we should select suitable heads for adversarial attacked images according to their adversarial strength. \ding{183} among the thirteen adversarial strengths, either $\text{head}\_3$ or $\text{head}\_4$ has the best performance. This inspires us to find the possibility of combining the advantage of heads for small attack strength and heads for large attack strength together to build a perturbation-aware filtering-based denoising method. In particular, as the Filt($L_\text{img}$) is only less effective on the median attack strengths and $\text{head}\_4$ is effective on these attack strengths, we may try to combine them together.

\begin{figure*}[tbp]
\centering
\includegraphics[width=0.95\linewidth]{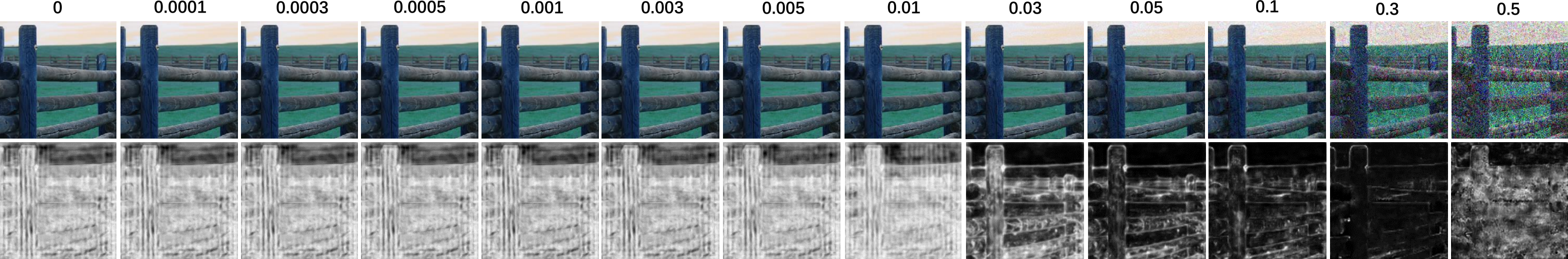}
\caption{The images in the first row are the noise images with different PGD attack strengths on ResNet50 and upon the same clean image selected from NeurIPS-2017DEV dataset. The values of the attack strength (\ie, $\epsilon$) are above the images. The images in the second row are the uncertainty maps corresponding to the above images. We can find that noise images with different attack strengths lead to different uncertainty maps.}
\label{fig:uncertainty_map}
\end{figure*}

\begin{SCtable*}[]
\scriptsize
\centering
\caption{Accuracy of the standard trained ResNet50 on adversarial examples of NeurIPS-2017DEV dataset (\ie, Acc. after Attack), and their denoised counterparts with predictive pertubation-aware denoising method.}
\setlength{\tabcolsep}{3pt}
\begin{tabular}{lccccccccccccc}
\hline 
& \multicolumn{12}{c}{$\text{PGD}(n=40,\epsilon)$}\tabularnewline
& 0 & $1\mathrm{e}^{-4}$ & $3\mathrm{e}^{-4}$ & $5\mathrm{e}^{-4}$ & $1\mathrm{e}^{-3}$ & $3\mathrm{e}^{-3}$ & $5\mathrm{e}^{-3}$ & $1\mathrm{e}^{-2}$ & $3\mathrm{e}^{-2}$ & $5\mathrm{e}^{-2}$ & $1\mathrm{e}^{-1}$ & $3\mathrm{e}^{-1}$ & $5\mathrm{e}^{-1}$ \tabularnewline
\midrule
Acc. After Attack & 0.913 & 0.886 & 0.733 & 0.605 & 0.268 & 0.018 & 0.006 & 0 & 0 & 0 & 0 & 0 & 0\tabularnewline
\midrule
PA-Filt($L_\text{img}$)$_\text{sl}$ & 0.934 & 0.903 & 0.852 & 0.772 & 0.532 & 0.062 & 0.015 & 0.006 & 0.036 & 0.071 & 0.047 & 0.277 & 0.037\tabularnewline
PA-Filt($L_\text{img}$)$_\text{m}$ & 0.875 & 0.869 & 0.863 & 0.852 & 0.817 & 0.658 & 0.505 & 0.285 & 0.159 & 0.126 & 0.059 & 0.247 & 0.042\tabularnewline
\hline 
PA-Filt($L_\text{img}$) & 0.923 & \textbf{0.912} & \textbf{0.883} & \textbf{0.859} & \textbf{0.768} & \textbf{0.295} & \textbf{0.118} & \textbf{0.034} & \textbf{0.074} & \textbf{0.091} & 0.045 & 0.269 & \textbf{0.037}\tabularnewline
Filt($L_\text{img}$) & 0.928 & 0.899 & 0.829 & 0.747 & 0.484 & 0.045 & 0.014 & 0.004 & 0.031 & 0.076 & 0.108 & 0.278 & 0.036\tabularnewline
\hline 
\end{tabular}
\label{Table:Pertubation-aware-network}
\end{SCtable*}

\section{Methodology}\label{sec:method}


\subsection{Predictive Perturbation-aware Filtering}
As shown in Fig.~\ref{fig:archs}, our method mainly contain two improvements to the filtering-based pixel denoising. We first propose \textit{dual-perturbation filtering} to obtain two types of filters. Each type of filters has advantages in dealing with specific attack strengths. For example, the first one does better on small \& large (\textbf{sl}) attack strengths while the second one does better on median (\textbf{m}) attack strengths. The \textit{dual-perturbation filtering} gives two outputs: $\tilde{\mathbf{I}}_\text{sl}$ and $\tilde{\mathbf{I}}_\text{m}$. \textbf{Please note} that the subscript `sl' and `m' mean that the filters are strong in dealing with (small \& large) and median attack strengths respectively, not mean the attack strengths of the adversarial images used to train the model. Second, we propose \textit{uncertainty-aware fusion} to combine $\tilde{\mathbf{I}}_\text{sl}$ and $\tilde{\mathbf{I}}_\text{m}$. The inputs of the \textit{uncertainty-aware fusion} are uncertainty maps (\ie, $\mathcal{U}_\text{sl}(\hat{\mathbf{I}})$ and $\mathcal{U}_\text{m}(\hat{\mathbf{I}})$) generated by the two filters and the output is a weight map. The weight map is used to fuse the previous two outputs together. 

\begin{figure}[tbp]
\centering
\includegraphics[width=0.9\linewidth]{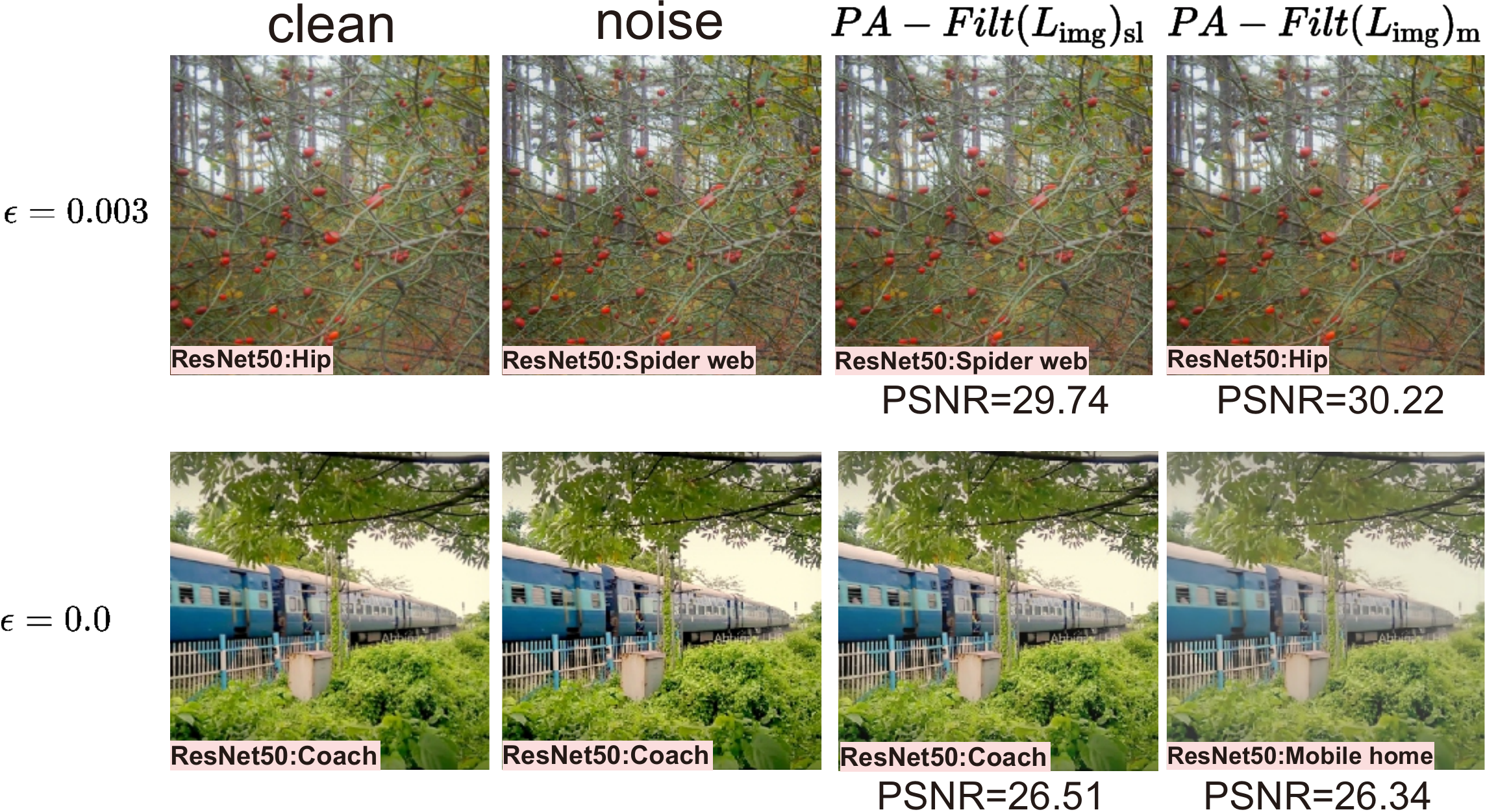}
\caption{The images in turn are clean images, noise images, output of PA-Filt($L_\text{img}$)$_\text{sl}$ (\ie, $\tilde{\mathbf{I}}_\text{sl}$) and output of PA-Filt($L_\text{img}$)$_\text{m}$ (\ie, $\tilde{\mathbf{I}}_\text{m}$). The noise image in the first row is attacked by PGD with $\epsilon=0.003$ while that in the second row is attacked by PGD with $\epsilon=0$ (\ie, the noise image is the same as clean image).}
\label{fig:output_dir_heavy}
\end{figure}

{\bf Dual-perturbation filtering.} Due to the second conclusion in Sec.~\ref{subsec:multi_head_filtering}, we need to design a network that can output two kinds of kernels. The first one should be the same with the Filt($L_\text{img}$) trained by all the attack strengths. The other one should be similar with Filt($L_\text{img}^*$) trained by large attack strength, which makes up for the defect of Filt($L_\text{img}$) in the median attack strength. To this end, we build a novel network that contains one encoder and two decoders. One decoder is designed for addressing the small and large adversarial noise while the second one is for the median strength. We denote this network as the Y-Net and the output of the two decoders as $\varphi_\text{sl}(\cdot)$ and $\varphi_\text{m}(\cdot)$, respectively.
Given an adversarial example $\hat{\mathbf{I}}$ and the denoising operation $\mathcal{D}(\cdot)$, the Y-Net produces two outputs based on the predictive noise information as 
%
\begin{align}\label{eq:denoised_dir}
\tilde{\mathbf{I}}_\text{sl} = \mathcal{D}(\varphi_\text{sl}(\hat{\mathbf{I}}),\hat{\mathbf{I}}),\quad\quad \tilde{\mathbf{I}}_\text{m} = \mathcal{D}(\varphi_\text{m}(\hat{\mathbf{I}}),\hat{\mathbf{I}}).
\end{align}
We denote above denoising method as `PA-Filt' as an extension of Filt. Meanwhile, the PA-Filt($L_\text{img}$)$_\text{sl}$ and PA-Filt($L_\text{img}$)$_\text{m}$ denote the denoisers with the two decoders, respectively.
As shown in Fig.~\ref{fig:output_dir_heavy}, the denoising results of PA-Filt($L_\text{img}$)$_\text{sl}$ and PA-Filt($L_\text{img}$)$_\text{m}$ (\ie, $\tilde{\mathbf{I}}_\text{sl}$ and $\tilde{\mathbf{I}}_\text{m}$) have different performance on various attack strengths.
Specifically, the noise image in the first row is attacked by PGD with $\epsilon=0.003$. We can find that $\tilde{\mathbf{I}}_\text{m}$ not only obtains higher PSNR than $\tilde{\mathbf{I}}_\text{sl}$, but also successfully defense the attack. On the contrary, $\tilde{\mathbf{I}}_\text{sl}$ fails to defense the attack. The conclusion is opposite with the example of $\epsilon=0.0$. In the second row, the noise image is attacked by PGD with $\epsilon=0.0$ (\ie, noise image is the same as clean image). Here the $\tilde{\mathbf{I}}_\text{sl}$ does better than $\tilde{\mathbf{I}}_\text{m}$. Dues to the quite different performance of $\tilde{\mathbf{I}}_\text{sl}$ and $\tilde{\mathbf{I}}_\text{m}$ on adversarial examples with various attack strength, we propose to fuse them together to overcome their shortages.

\begin{figure}[tbp]
\centering
\includegraphics[width=0.95\linewidth]{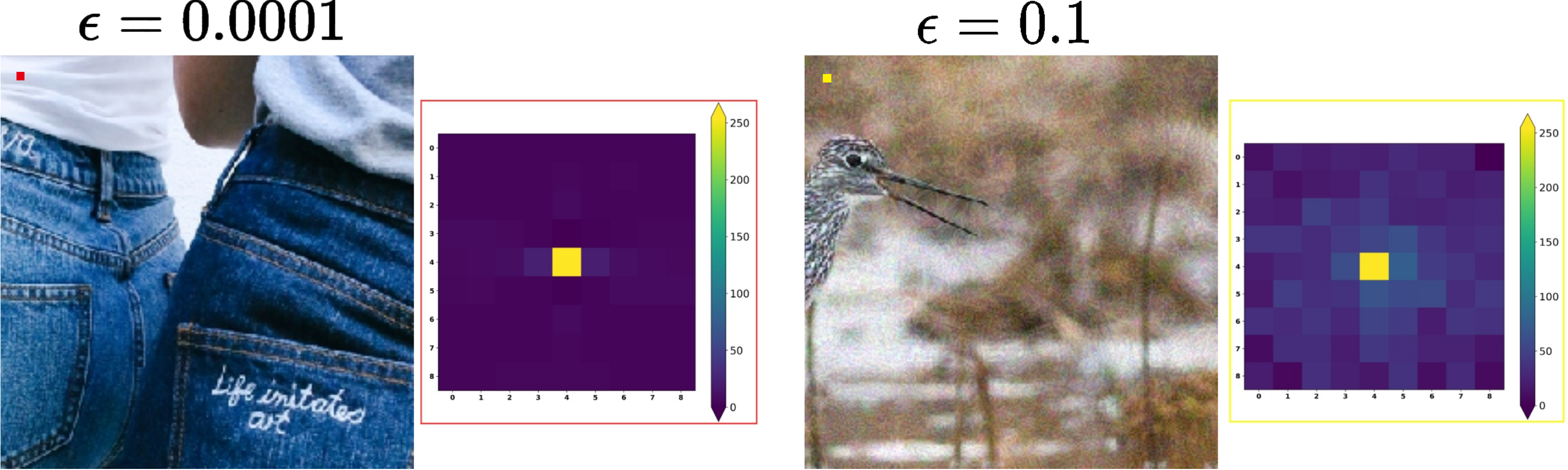}
\caption{Here shows the kernels of $\varphi_\text{sl}(\cdot)$ for noise images with two different attack strengths. The left pair shows a noise image with $\epsilon=0.0001$ and the kernel for filtering the pixel that label in red (\ie, location (12,12)). The right pair shows a noise image. The $\epsilon$ is 0.1 and the kernel for filtering the pixel that label in yellow (\ie, location (12,12)) of the noise image is on the right.}
\label{fig:kernel_demo}
\end{figure}

{\bf Uncertainty-aware fusion.} 
Based on the $\tilde{\mathbf{I}}_\text{sl}$ and $\tilde{\mathbf{I}}_\text{m}$, we urgently need a weight map to fuse them together. 

As introduced before in Sec.~\ref{subsec:formulations}, each pixel of $\hat{\mathbf{I}}$ is filtered by an exclusive kernel estimated from $\varphi(\hat{\mathbf{I}})\in\mathbb{R}^{H\times W\times 3K^2}$. Here, we present that the kernels' patterns are related to different attack strengths. 
Specifically, in Fig.~\ref{fig:kernel_demo}, we show the kernels of $\varphi_\text{sl}(\cdot)$ for noise images with two different attack strengths. 
The two subfigures show two noise images with two kernels of specified pixels (\ie, the red point and yellow point) under $\epsilon=0.0001$ and $\epsilon=0.1$, respectively.
We observe that for the noise image with small attack strength, the kernel almost does not change the pixel while for the noise image with strong attack strength, the kernel changes the pixel a lot. Intuitively, we can say that the kernels for all the pixels in the noise image are strongly related to the attack strengths.
As a result, We could fuse the $\tilde{\mathbf{I}}_\text{sl}$ and $\tilde{\mathbf{I}}_\text{m}$ under the guidance of the predicted kernels. Here, we propose to build an \textit{uncertainty map} based on the pixel-wise kernels. 

As introduced in Sec.~\ref{subsec:formulations}, each pixel of $\hat{\mathbf{I}}$ is filtered by an exclusive kernel estimated from $\varphi(\hat{\mathbf{I}})\in\mathbb{R}^{H\times W\times 3K^2}$. 
The kernel for pixel in the $m$th row and $n$th column ($0\leq m \leq H, 0\leq n \leq W$) of the noise image $\hat{\mathbf{I}}$ is $\varphi(\hat{\mathbf{I}})_{m,n}\in\mathbb{R}^{3K^2}$. The uncertainty map defined by us is $\mathcal{U}\in\mathbb{R}^{H\times W}$. The pixel $\mathcal{U}_{m,n}$ of uncertainty map $\mathcal{U}$ for noise image $\hat{\mathbf{I}}$ is calculated by a max pooling layer
\begin{align}\label{eq:uncertainty_map}
\mathcal{U}(\hat{\mathbf{I}})_{m,n} = \text{maxpool}(\varphi(\hat{\mathbf{I}})_{m,n}).
\end{align}
%
We use max-pooling to calculate the uncertainty map since the results can reflect the strength of the noise.
As shown in Fig.~\ref{fig:uncertainty_map}, we show the noise images of different attack strengths and their corresponding uncertainty maps generated by \textit{max-pooling} based on model Filt($L_\text{img}$). We can find that the kernels for all the pixels in the noise image are strongly related to the attack strengths.

According to above formulation, we can get two uncertainty maps, \ie, $\mathcal{U}_\text{sl}(\hat{\mathbf{I}})$ and $\mathcal{U}_\text{m}(\hat{\mathbf{I}})$, by using  $\varphi_\text{sl}(\hat{\mathbf{I}})$ and $\varphi_\text{m}(\hat{\mathbf{I}})$ for Eq.~\eqref{eq:uncertainty_map}, respectively. 
Then, we concatenate the maps together and use a convolutional network $C(\cdot)$ to generate a weight map $\mathcal{W}\in\mathbb{R}^{H\times W}$ and denote the weight map for noise image $\hat{\mathbf{I}}$ as $\mathcal{W}(\hat{\mathbf{I}})$
\begin{align}\label{eq:weight_map}
\mathcal{W}(\hat{\mathbf{I}}) = C([\mathcal{U}_\text{sl}(\hat{\mathbf{I}}),\mathcal{U}_\text{m}(\hat{\mathbf{I}})]).
\end{align}
The final output $\tilde{\mathbf{I}}$ of the entire network is 
\begin{align}\label{eq:fusion}
\tilde{\mathbf{I}} = \mathcal{W}(\hat{\mathbf{I}}) \odot \tilde{\mathbf{I}}_\text{sl} + (1-\mathcal{W}(\hat{\mathbf{I}})) \odot \tilde{\mathbf{I}}_\text{m},
\end{align}
where $\odot$ means element-wise multiplication.
We show the performance of our method in Table~\ref{Table:Pertubation-aware-network} where `PA-Filt($L_\text{img}$)$_\text{sl}$' means the output image of $\varphi_\text{sl}(\cdot)$ while `PA-Filt($L_\text{img}$)$_\text{m}$' means the output image of $\varphi_\text{m}(\cdot)$. We can find that they have obvious advantages and disadvantages under different attack strengths. `PA-Filt($L_\text{img}$)' means the output of our entire network, which significantly outperforms the Filt($L_\text{img}$) on most of the attack strengths (in bold).

\begin{SCtable*}
\scriptsize
\centering
\caption{Accuracy of the adversarially trained ResNet50 on adversarial examples of ImageNet (\ie, Acc. after Attack), and their denoised counterparts with different denoising methods.}
\setlength{\tabcolsep}{2.4pt}
\begin{tabular}{cccccccccccccc|ccccc}
\hline 
& \multicolumn{13}{c|}{$\text{PGD}(n=40,\epsilon)$} & \multicolumn{5}{c}{ $\text{PGD}(n,\epsilon=0.3)$}\tabularnewline
& 0 & $1\mathrm{e}^{-4}$ & $3\mathrm{e}^{-4}$ & $5\mathrm{e}^{-4}$ & $1\mathrm{e}^{-3}$ & $3\mathrm{e}^{-3}$ & $5\mathrm{e}^{-3}$ & $1\mathrm{e}^{-2}$ & $3\mathrm{e}^{-2}$ & $5\mathrm{e}^{-2}$ & $1\mathrm{e}^{-1}$ & $3\mathrm{e}^{-1}$ & $5\mathrm{e}^{-1}$ & 10 & 30 & 50 & 70 & 90\tabularnewline
\midrule
Acc. after Attack & 0.913 & 0.866 & 0.733 & 0.605 & 0.268 & 0.018 & 0.006 & 0 & 0 & 0 & 0 & 0 & 0 & 0 & 0 & 0 & 0 & 0\tabularnewline
\midrule
Adv & 0.628 & 0.628 & 0.628 & 0.628 & 0.628 & 0.627 & 0.626 & 0.625 & 0.624 & 0.607 & 0.558 & 0.148 & 0.039 & 0.132 & 0.159 & 0.169 & 0.175 & 0.178\tabularnewline
Add+Adv & 0.630 & 0.630 & 0.630 & 0.630 & 0.629 & 0.628 & 0.626 & 0.621 & 0.621 & 0.609 & 0.571 & 0.290 & 0.092 & 0.275 & 0.287 & 0.295 & 0.302 & 0.299\tabularnewline
Filt+Adv & 0.626 & 0.626 & 0.626 & 0.626 & 0.626 & 0.625 & 0.625 & 0.622 & 0.613 & 0.610 & 0.590 & 0.455 & 0.341 & 0.452 & 0.449 & 0.455 & 0.456 & 0.457\tabularnewline 
PA-Filt+Adv & 0.614 & 0.614 & 0.614 &0.615 &0.614 & 0.612 & 0.611 & 0.610 & 0.609 & 0.605 & 0.592 & 0.452 & 0.345 & 0.445& 0.442& 0.452& 0.446& 0.447
\tabularnewline
\hline 
\end{tabular}
\label{Table:Improve_adversarial_resnet50_model}
\end{SCtable*}

\begin{SCtable*}
\scriptsize
\centering
\caption{Accuracy of the adversarially trained ResNet50 (\ie, FPD) on adversarial examples of SVHN (\ie, Acc. after Attack), and their denoised counterparts with FPD and enhanced FPD.}
\setlength{\tabcolsep}{5pt}
\begin{tabular}{cccccccccccccc}
\hline 
& \multicolumn{13}{c}{$\text{PGD}(n=40,\epsilon)$} \tabularnewline
& 0 & $1\mathrm{e}^{-4}$ & $3\mathrm{e}^{-4}$ & $5\mathrm{e}^{-4}$ & $1\mathrm{e}^{-3}$ & $3\mathrm{e}^{-3}$ & $5\mathrm{e}^{-3}$ & $1\mathrm{e}^{-2}$ & $3\mathrm{e}^{-2}$ & $5\mathrm{e}^{-2}$ & $1\mathrm{e}^{-1}$ & $3\mathrm{e}^{-1}$ & $5\mathrm{e}^{-1}$ \tabularnewline
\midrule
Acc. after Attack & 0.925 & 0.936 & 0.905 & 0.918 & 0.891 & 0.781 & 0.647 & 0.338 & 0.040 & 0.008 & 0.005 & 0.001 & 0 \tabularnewline
\midrule
FPD & 0.881 & 0.888 & 0.879 & 0.868 & 0.881 & 0.871 & 0.878 & 0.856 & 0.795 & 0.793 & 0.700 & 0.534 & 0.335 \tabularnewline
Add+FPD & 0.879 & 0.887 & 0.881 & 0.869 & 0.883 & 0.874 & 0.890 & 0.861 & 0.834 & 0.857 & 0.835 & 0.777 & 0.567 \tabularnewline
Filt+FPD & 0.875 & 0.872 & 0.874 & 0.872 & 0.865 & 0.863 & 0.872 & 0.850 & 0.792 & 0.799 & 0.710 & 0.570 & 0.373

\tabularnewline
\hline 
\end{tabular}
\label{Table:Improve_FPD}
\end{SCtable*}

\subsection{Multi-domain Learning}

The proposed \textit{dual-perturbation filtering} contains two decoders for adversarial noise recovery with different noise strengths. We can regard the two decoders as domain-specific architectures, which takes the charge of handling noise with (small \& large) and median strengths, respectively. Intuitively, we can update the parameters of the two decoders independently according to the input adversarial examples' noise strength during the training process. As a result, the two decoders will adapt to different noise domains. To better understand the whole process, we detail the architecture and training details in the following. 

{\bf Architecture.} 
The Y-Net of our model has the same encoder and two same decoders as that of the U-Net in the \textbf{Architecture of $\varphi(\cdot)$} of Sec.~\ref{subsec:setups_empirical}. The convolutional network $C(\cdot)$ which used by us to generate a weight map from uncertainty maps contains six convolutional layers followed by a Sigmoid function.

{\bf Training details.} 
We first train the Y-Net with $L_1$ loss function. The decoder of $\varphi_\text{sl}(\cdot)$ is updated with 12 different adversarial strength, \ie, $\epsilon\in\{1\mathrm{e}^{-m}, 3\mathrm{e}^{-m}, 5\mathrm{e}^{-m}|m\in[1,2,3,4]\}$ while the decoder of $\varphi_\text{m}(\cdot)$ is updated with $\epsilon\in\{1\mathrm{e}^{-1}, 3\mathrm{e}^{-1}, 5\mathrm{e}^{-1}\}$. Then we fixed the parameters of the Y-Net and use 12 different adversarial strength, \ie, $\epsilon\in\{1\mathrm{e}^{-m}, 3\mathrm{e}^{-m}, 5\mathrm{e}^{-m}|m\in[1,2,3,4]\}$ to train \textit{uncertainty-aware fusion} module. As the $\varphi_\text{sl}(\cdot)$ and $\varphi_\text{m}(\cdot)$ have been trained by $L_1$ loss, to improve the classification performance, we use semantic-level loss to train \textit{uncertainty-aware fusion}. We take the classification output of ResNet50 with the denoised image and the clean image as input to calculate cross-entropy loss. The adversarial attack method and training dataset are the same as in Sec.~\ref{subsec:setups_empirical}.

\section{Adversarial Denoising \& Training}
Pixel-wise denoising focuses on enhancing the quality of data while adversarial training allows the network to see more difficult data, especially the attacked images. These two methods deal with the defense task from two reasonable angles, which inspires us to think about whether the classification result would be better or not by combining them together. In this section, we formulate the method mathematically and detail the experiments in Sec.~\ref{sec:comparison_experiment}.

Instead of training a standard trained CNN (\ie, $\phi(\cdot)$) by minimizing the loss function, we can improve its adversarial robustness by conducting the adversarial training, that is, to produce the adversarial examples during the training procedure and solve a min-max objective function 
\begin{align}
\min_{\theta}\mathbb{E}_{(\mathbf{I},y)\sim D}&(\max_{\delta^{*}}\mathcal{J}(\phi')(\mathbf{I}+\delta^{*}),y) \text{~~subject to}~\|\delta^{*}\|_p\leq\epsilon, \label{eq:adversarial_training}
\end{align}
where $\mathcal{J}(\cdot)$ denotes the loss function with $y$ being the annotation of the input image $\mathbf{I}$. $\phi'$ represents the adversarially trained model and $\theta$ denotes the model parameters.

During the evaluation process, we propose to conduct the adversarial denoising on the adversarial examples first and feed denoising results to the adversarially trained model $\phi'$.
Specifically, given an imperceptible adversarial example $\hat{\mathbf{I}}$, we can use denoising operation $\mathcal{D}(\cdot)$ to obtain the denoised image $\tilde{\mathbf{I}}$ first, as shown in Eq.~(\ref{eq:unifieddenoiser}). Then, we use an adversarially trained model $\phi'(\cdot)$ to classify the denoised image $\tilde{\mathbf{I}}$ and predict the category of the image.

\section{Experiments}\label{sec:exp}
\textbf{Defense methods and datasets.}
For adversarial training methods, feature pyramid decoder (FPD) \cite{li2020enhancing} and feature denoising \cite{xie2019feature} (FD) are the two state-of-the-art methods that we investigate. We use ImageNet \cite{deng2009imagenet}, NeurIPS-2017DEV \cite{kurakin2018adversarial}, and SVHN \cite{svhn} as the training and test datasets. ImageNet is a well-known 1000-class dataset. SVHN is a real-world colored digits image dataset. It includes 73,257 MNIST-like 32 $\times$ 32 images centered around a single character. NeurIPS-2017DEV is introduced in the testing dataset of Sec.~\ref{subsec:setups_empirical}.

\textbf{Evaluation settings.}
For experiments that take ImageNet as the training dataset, we use the same setup as in Sec.~\ref{subsec:setups_empirical}.
For experiments that take SVHN as the training dataset, we randomly use 10,000 images of SVHN to train the model. We also add 12 different intensities \ie, $\epsilon\in\{1\mathrm{e}^{-m}, 3\mathrm{e}^{-m}, 5\mathrm{e}^{-m}|m\in[1,2,3,4]\}$ of $L_{\infty}$-PGD noises as before to expand the training dataset to 120,000 images.

The test datasets of us mainly consist of NeurIPS-2017DEV and SVHN. We also use Foolbox \cite{rauber2017foolbox} to add adversarial noise. Whether to use a different test dataset is based on the baseline method we compared with. Each test dataset has 1,000 images and is expanded to 12,000 images by 12 different $\epsilon$ as before.

\textbf{Metrics.}
The main metric is the classification accuracy of the methods. We compare the classification accuracy of our method and baseline methods on different adversarial noise and $\epsilon$. 

All the experiments are carried out on a Ubuntu 16.04 system with an Intel(R) Xeon(R) CPU E5-2699 with 196 GB of RAM, equipped with one Tesla V100 GPU of 32G RAM. 

\begin{SCtable*}
\scriptsize
\centering
\caption{Accuracy of the adversarially trained ResNeXt101 (\ie, FD) on adversarial examples of NeurIPS-2017DEV (\ie, Acc. after Attack), and their denoised counterparts with FD and enhanced FD.}
\setlength{\tabcolsep}{5pt}
\begin{tabular}{cccccccccccccc}
\hline 
 & \multicolumn{13}{c}{$\text{PGD}(n=40,\epsilon)$} \tabularnewline
 & 0 & $1\mathrm{e}^{-4}$ & $3\mathrm{e}^{-4}$ & $5\mathrm{e}^{-4}$ & $1\mathrm{e}^{-3}$ & $3\mathrm{e}^{-3}$ & $5\mathrm{e}^{-3}$ & $1\mathrm{e}^{-2}$ & $3\mathrm{e}^{-2}$ & $5\mathrm{e}^{-2}$ & $1\mathrm{e}^{-1}$ & $3\mathrm{e}^{-1}$ & $5\mathrm{e}^{-1}$ \tabularnewline
\midrule 
Acc. after Attack & 0.935 & 0.895 & 0.785 & 0.631 & 0.333 & 0.039 & 0.015 & 0.005 & 0.004 & 0.002 & 0.003 & 0.002 & 0.001 \tabularnewline
\midrule
FD & 0.820 & 0.820 & 0.820  & 0.820 & 0.820 & 0.820 & 0.818 & 0.816 & 0.814 & 0.814 & 0.800 & 0.427 & 0.108 \tabularnewline
Add+FD & 0.812 & 0.811 & 0.811 & 0.810 & 0.809 & 0.810 & 0.810 & 0.811 & 0.814 & 0.810 & 0.791 & 0.623 & 0.266 \tabularnewline
Filt+FD & 0.818 & 0.818 & 0.819 & 0.818 & 0.818 & 0.819 & 0.819 & 0.817 & 0.812 & 0.802 & 0.790 & 0.692 & 0.619
\tabularnewline
\hline 
\end{tabular}
\label{Table:Improve_FD}
\end{SCtable*}

\subsection{Comparison Experiments}\label{sec:comparison_experiment}
There is an interesting question that whether combining adversarial denoising and adversarial training defense together will obtain more robust classification results. The following subsections provide comprehensive experiments to answer this question.

\subsubsection{Improvement for basic adversarially trained model}
First, we present the fundamental experiment on demonstrating the combination of basic adversarially trained ResNet50 model with additive-based and filtering-based denoising respectively. As shown in Table~\ref{Table:Improve_adversarial_resnet50_model}, There are four types of denoising groups in the first column. `Adv' represents the adversarially trained ResNet50 model from the open-sourced resource of \cite{xie2020smooth}. 

`Add+Adv' and `Filt+Adv' refer to using the additive-based and filtering-based method respectively to the denoise image first and then using adversarially trained ResNet50 model. From the table we can find that the adversarially trained ResNet50 model has poor performance on small attack strengths while having good performance on large attack strengths. When using additive-based and filtering-based denoising methods to enhance the adversarially trained ResNet50 model, significant improvement exists on the images with large attack strength. 

`PA-Filt+Adv' refers to using our predictive perturbation-aware filtering-based method to denoise image first and then using adversarially trained ResNet50 model. The performance on improving the adversarially trained model is similar to that of `Filt+Adv'. We also build a test dataset that keeps the attack strength constant (\ie, 0.3) and changes the attack iterations (\eg, 10, 30, 50, 70, 90). We can observe that the conclusion is the same as before and the performance of all the four groups is stable. The above experiments demonstrate that the adversarial denoising defense could benefit the adversarial training models, which encourages us to achieve more robust image classification results by considering data and model enhancement methods jointly.

\subsubsection{Improvement for FPD}

To verify whether the conclusion is the same on the state-of-the-art adversarial training methods, we conduct further experiments on them.

Feature pyramid decoder (FPD) \cite{li2020enhancing} is the state-of-the-art adversarial training method. They demonstrate that FPD-enhanced CNNs gain sufficient robustness against general adversarial samples on SVHN. To test the performance of FPD on various noise images with different perturbation amplitude, and how can pixel denoising method benefit them, we build a testing dataset with respect to the model provided by FPD. FPD provides an adversarially trained ResNet50 on dataset SVHN. Thus we train a ResNet50 on the SVHN training dataset and attack the model by PGD to generate an adversarial attacked SVHN testing dataset. The accuracy of adversarial samples is shown in the second row of Table~\ref{Table:Improve_FPD}. 

In the first column, `FPD' represents directly using their model to obtain the classification accuracy. We can see that FPD does have a good result when the $\epsilon$ is larger than 0.003. The pixel denoising models used to enhance FPD are trained by using the PGD-attacked images on ResNet50 with L1 loss function. `Filt+FPD' means using filtering-based denoising to process the noise image and then classified by FPD. `Add+FPD' means using additive-based denoising before using FPD. We can find that no matter using filtering-based denoising and additive-based denoising before FPD, the accuracy will increase when the $\epsilon$ is bigger than 0.05 while maintaining a similar accuracy with FPD when the $\epsilon$ is smaller than 0.05. `Add+FPD' has especially better performance when the $\epsilon$ is bigger than 0.05.

Through the observation that FPD can be benefited by the pre-processing of filtering-based and additive-based denoising, we can reach a preliminary conclusion that the robustness of the state-of-the-art adversarial training models can be enhanced by pixel denoising method to some extent.

\subsubsection{Improvement for FD}

Feature denoising (FD) \cite{xie2019feature} is also a state-of-the-art adversarial training method. It has achieved the championship of the Competition on Adversarial Attacks and Defenses (CAAD). Due to the fact that adversarial perturbations on images lead to noise in the features constructed by these networks, they develop new network architectures that increase adversarial robustness by performing feature denoising. Their model is based on ResNeXt101 \cite{xie2017aggregated} and shows excellent robustness against PGD attacks on the ImageNet-like NeurIPS-2017DEV dataset. 

The test dataset is obtained by attacking the NeurIPS-2017DEV dataset with different perturbation amplitudes. FD provided an adversarially trained ResNeXt101 on dataset ImageNet. Thus we use the pre-trained ResNeXt101 model provided by PyTorch and attack the model by PGD to generate an adversarial attacked NeurIPS-2017DEV testing dataset. The accuracy of adversarial samples is shown in the second row of Table~\ref{Table:Improve_FD}. 

In the first column, `FD' represents direct using their model to achieve the classification accuracy. We can find that FD does well on the test dataset when the $\epsilon$ $\textgreater$ 0.0003. We use PGD-attacked images on ResNeXt101 as the training dataset to train our pixel denoising models. The loss function used in the training procedure is L1 loss. `Filt+FD' means using filtering-based denoising to process the noise image and then classified by FD. `Add+FD' means using additive-based denoising before using FD. We can find that although FD has excellent accuracy on classifying images on large $\epsilon$, filtering-based and additive-based denoising methods can both improve the accuracy. In particular, the filtering-based method can increase the accuracy by 5.7 times of FD when $\epsilon$ is 0.5.

The improvement in the accuracy of FD by adversarial denoising methods further demonstrates the complementary of adversarial denoising and training methods.

\vspace{-10pt}
\section{Conclusion}\label{sec:concl}
In this paper, we reveal the better performance of the filtering-based pixel denoising method than the additive-based pixel denoising method on adversarial defense. Based on the comparison, we observe the flaw of the filtering-based denoising method in dealing with adversarial examples with various perturbation amplitudes. Thus we propose a predictive perturbation-aware \& pixel-wise filtering method to automatically denoise the image. Furthermore, we also expose the complementarity of adversarial denoising defense and adversarial training defense, which encourages further research on this combination domain.

Going forward, it is worth investigating beyond traditional additive noise-based adversarial attacks. Recent studies on non additive noise-based adversarial attacks have caught attention such as adversarial weather elements, \eg, rain \cite{arxiv20_advrain} and haze \cite{gao2021advhaze}, and image degradation-mimetic adversarial attacks \eg, adversarial exposure \cite{icme21_xray,arxiv20_retinopathy}, vignetting \cite{ijcai21_ava}, blur \cite{neurips20_abba}, color jittering \cite{arxiv20_cosal}, \etc, and how can we extend the proposed AdvFilter to accommodate these scenarios is also an important research direction.

\textbf{ACKNOWLEDGMENTS} This work is supported by National Key Research and Development Program (2020AAA0107800). It is also supported by the NSFCs of China No. 61872144, the National Research Foundation, Singapore under its the AI Singapore Programme (AISG2-RP-2020-019), the National Research Foundation, Prime Ministers Office, Singapore under its National Cybersecurity R\&D Program (No. NRF2018NCR-NCR005-0001), NRF Investigatorship NRFI06-2020-0001, the National Research Foundation through its National Satellite of Excellence in Trustworthy Software Systems (NSOE-TSS) project under the National Cybersecurity R\&D (NCR) Grant (No.~NRF2018NCR-NSOE003-0001). We gratefully acknowledge the support of NVIDIA AI Tech Center (NVAITC).

\bibliographystyle{ACM-Reference-Format}
\balance 
\bibliography{ref}


\end{document}